\documentclass[sigconf]{acmart}
\usepackage[utf8]{inputenc}
\usepackage{amsmath,amsfonts}
\usepackage{algorithmic}
\usepackage{algorithm}
\usepackage{array}
\usepackage[caption=false,font=normalsize,labelfont=sf,textfont=sf]{subfig}
\usepackage{textcomp}
\usepackage{stfloats}
\usepackage{url}
\usepackage{verbatim}
\usepackage{graphicx}
\usepackage{cite}
\usepackage{tabularx}
\usepackage{hyperref}
\usepackage{hyperref}
\hypersetup{
    colorlinks=true,
    linkcolor=blue,
    filecolor=magenta,      
    urlcolor=cyan,
}
\usepackage{outlines}
\usepackage{dsfont}
\usepackage{xcolor}
\usepackage{cleveref}

\usepackage{float}
\usepackage{caption}
\usepackage{listings}
\usepackage{simplebnf}
\usepackage{multirow}
\usepackage{threeparttable}

\newfloat{lstfloat}{t}{lop}
\floatname{lstfloat}{listing}

\usepackage[acronym, toc, nonumberlist]{glossaries}
\newacronym{snv}{SNV}{Single Nucleotide Variant}
\newacronym{iad}{IAD}{Increase Archive Diversity}
\newacronym{psi}{PSI}{percent spliced-in}
\newacronym{gp}{GP}{Genetic Programming}
\newacronym{gggp}{GGGP}{Grammar-Guided Genetic Programming}
\newacronym{tpe}{TPE}{Tree-structured Parzen Estimator Approach}
\newacronym{rs}{RS}{Random Search}
\newacronym{kd}{KD}{Knockdown}
\newacronym{rnaseq}{RNA-Seq}{RNA Sequencing}
\newacronym{dress}{DRESS}{Deep learning based Resource for Exploring Splicing Signatures}
\newacronym{bnf}{BNF}{Backus Normal Form}
\newacronym{gan}{GANs}{Generative adversarial networks}
\newacronym{vae}{VAEs}{Variational Autoencoders}
\newacronym{pwms}{PWMs}{Position Weight Matrices}
\newacronym{rbps}{RBPs}{RNA-binding proteins}

\copyrightyear{2024}
\acmYear{2024}
\setcopyright{rightsretained}
\acmConference[GECCO '24]{Genetic and Evolutionary Computation
Conference}{July 14--18, 2024}{Melbourne, VIC, Australia}
\acmBooktitle{Genetic and Evolutionary Computation Conference (GECCO '24),
July 14-18, 2024, Melbourne, VIC, Australia}
\acmDOI{10.1145/3638529.3653990}
\acmISBN{979-8-4007-0494-9/24/07}

\begin{document}
\title{Semantically Rich Local Dataset Generation for Explainable AI in Genomics}

\author{Pedro Barbosa}
\affiliation{%
  \institution{LASIGE, Faculdade de Ciências da Universidade de Lisboa}
  \institution{Instituto de Medicina Molecular João Lobo Antunes, Faculdade de Medicina da Universidade de Lisboa}
  \country{Portugal}
}
\email{psbarbosa@ciencias.ulisboa.pt}
\orcid{0000-0002-3892-7640}

\author{Rosina Savisaar}
\affiliation{%
  \institution{Mondego Science}
  \country{France}
}
\email{rosina@mondegoscience.com}
\orcid{0000-0002-8367-2317}

\author{Alcides Fonseca}
\affiliation{%
  \institution{LASIGE, Faculdade de Ciências da Universidade de Lisboa}
  \country{Portugal}
}
\email{me@alcidesfonseca.com}
\orcid{1234-5678-9012}

\begin{abstract}

Black box deep learning models trained on genomic sequences excel at predicting the outcomes of different gene regulatory mechanisms. Therefore, interpreting these models may provide novel insights into the underlying biology, supporting downstream biomedical applications. Due to their complexity, interpretable surrogate models can only be built for local explanations (e.g., a single instance). However, accomplishing this requires generating a dataset in the neighborhood of the input, which must maintain syntactic similarity to the original data while introducing semantic variability in the model's predictions. This task is challenging due to the complex sequence-to-function relationship of DNA. 

We propose using Genetic Programming to generate datasets by evolving perturbations in sequences that contribute to their semantic diversity. Our custom, domain-guided individual representation effectively constrains syntactic similarity, and we provide two alternative fitness functions that promote diversity with no computational effort. Applied to the RNA splicing domain, our approach quickly achieves good diversity and significantly outperforms a random baseline in exploring the search space, as shown by our proof-of-concept, short RNA sequence. Furthermore, we assess its generalizability and demonstrate scalability to larger sequences, resulting in a $\approx$30\% improvement over the baseline.

\end{abstract}
    


\maketitle

\section{Introduction}
\label{sec:introduction}

Gene regulation mechanisms are extremely complex in a way that humans are still far from understanding. Deep learning has demonstrated remarkable performance across a wide range of genomics tasks, especially when only using raw sequences as input \citep{zhouPredictingEffectsNoncoding2015,alipanahiPredictingSequenceSpecificities2015,kelleyBassetLearningRegulatory2016,jaganathanPredictingSplicingPrimary2019,avsecBaseresolutionModelsTranscriptionfactor2021,avsecEffectiveGeneExpression2021,dealmeidaDeepSTARRPredictsEnhancer2022}. Since these models approximate the relationship between the sequence space and the resulting phenotype, they have been used as an oracle for expensive and time-consuming wet lab experiments \citep{vaishnavEvolutionEvolvabilityEngineering2022,novakovskyObtainingGeneticsInsights2022}.

However, understanding how these models arrive at a prediction remains a challenge due to the number of parameters of these deep neural networks, making them difficult to interpret \citep{chakrabortyInterpretabilityDeepLearning2017}. Despite the progress made in designing interpretable-by-design deep nets \citep{ghorbaniAutomaticConceptbasedExplanations2019,kohConceptBottleneckModels2020,novakovskyExplaiNNInterpretableTransparent2023}, \textit{post-hoc} interpretation techniques are the most common strategy to study what the model learns (via input perturbations \citep{zhouPredictingEffectsNoncoding2015,nairFastISMPerformantSilico2022} or attribution-based methods \citep{shrikumarLearningImportantFeatures2017, sundararajanAxiomaticAttributionDeep2017,jhaEnhancedIntegratedGradients2020}). These techniques reveal important patterns within the sequences but do not infer semantic rules of the regulatory grammar without additional analyses \citep{avsecBaseresolutionModelsTranscriptionfactor2021,dealmeidaDeepSTARRPredictsEnhancer2022}. 

Alternatively, using inherently interpretable surrogate models
to emulate the black box model is worthy of consideration. Although global surrogates have been proposed for explaining complex models in other domains \citep{wuSparsityTreeRegularization2017,evansWhatBlackboxGenetic2019}, applying them in genomics remains a significant challenge due to the intricate combinatorial complexity within regulatory DNA \citep{novakovskyObtainingGeneticsInsights2022}. A more practical strategy is to train surrogate models on smaller subsets of the data to achieve local explanations \citep{ribeiroWhyShouldTrust2016,guidottiLocalRuleBasedExplanations2018,ferreiraApplyingGeneticProgramming2020,seitzInterpretingCisRegulatory2023}. For such surrogates to succeed, it is important that the local datasets contain sequences that are not only syntactically similar but also semantically diverse, so as to extensively sample the fitness landscape. 

The generation of such a local dataset is a challenging task, considering the vast combinatorial search space and the irregular fitness landscape. It requires exploring the syntactic neighborhood of the target sequence and identifying perturbed sequences that maximize the diversity of the semantic space. The search space grows linearly with the length of the sequence and exponentially with the number of simultaneous single nucleotide perturbations. It grows even faster when considering more complex, but realistic perturbations, such as insertions of deletions of short lengths. As a result, techniques such as exhaustive search or random search may not be feasible in this context. Exhaustive search becomes computationally intractable, while random search is oblivious to the semantic space, which may lead to datasets that sparsely cover the fitness landscape (\Cref{sec:relatedwork}).

To tackle this problem, we propose using \gls{gp}~\citep{kozaGeneticProgrammingProgramming1992} with a custom domain-aware grammar that restricts the perturbations applied to the original sequence. We also define two fitness functions, \textit{Bin filler} and \textit{Increased Archive Diversity}, that assess the potential of each sequence to improve the quality of the dataset (\Cref{sec:approach}).

As a case study (\Cref{sec:evaluation_methodology}), we focus on RNA splicing, an essential biological process that occurs during gene expression. In splicing, RNA sequences are edited to remove certain regions (\textit{introns}), after which the remaining blocks (\textit{exons}) are joined together. The information that governs this process is encoded in a series of poorly defined regulatory signals with variable locations within the sequence and is not yet fully understood~\citep{rogalskaRegulationPremRNASplicing2022}. In particular, these regulatory signals can span thousands of nucleotides, highlighting the complex relationship between the genotype and the splicing outcome.

We explore different hyperparameters on a controlled sequence, concluding that \gls{gp} is significantly more effective than randomly sampling the search space (\Cref{subsec:results}). Additionally, we show that this difference extends to a diverse set of input sequences, underscoring the effectiveness of our approach in generating locally diverse datasets (\Cref{sec:conclusion}).

\section{Related work}
\label{sec:relatedwork}

We consider previous research explicitly focusing on (or partially addressing) synthetic data generation in genomics as related work. In particular, we explore various applications of sequence generation guided by performant deep learning models. 

One such application is sequence design, which involves designing molecules with desired outcomes, such as controlling gene expression levels  \citep{zrimecControllingGeneExpression2022} or developing more efficient mRNA vaccines \citep{castillo-hairMachineLearningDesigning2022}. To do this, deep generative networks are used to model distributions of sequences with desired properties \citep{killoranGeneratingDesigningDNA2017,linderGenerativeNeuralNetwork2020,brookesConditioningAdaptiveSampling2019,linderFastActivationMaximization2021,valeriBioAutoMATEDEndtoendAutomated2023}. These frameworks employ activation maximization to maximize the property of interest by gradient ascent through the model oracle, typically using an appropriate generative network based on \gls{gan} \citep{goodfellowGenerativeAdversarialNetworks2014} or \gls{vae} \citep{kingmaAutoEncodingVariationalBayes2022}. Genetic algorithms have also been proposed, either by greedily perturbing the best sequences \citep{sinaiAdaLeadSimpleRobust2020}, employing very large population sizes \citep{angermuellerPopulationBasedBlackBoxOptimization2020} or evolving ensembles of optimization algorithms \citep{angermuellerPopulationBasedBlackBoxOptimization2020}. Recently, inspired by classical wet lab experiments, directed sequence evolution has been used successfully to iteratively evolve a random sequence into a synthetic, biologically functional sequence~\citep{kliePredictiveAnalysesRegulatory2023,taskiranCellTypeDirected2023}. This method exhaustively measures the impact of \gls{snv}s at each iteration and selects the perturbation that yields the largest model prediction change to be applied to the sequence. However, it's important to note that despite their utility, these approaches have inherent constraints. They do not generate sequences covering the entire fitness landscape of the oracle (the semantic space). In addition, many methods were not developed for model explainability, and can only scale to short sequence contexts.

The use of synthetic data augmentations has been proposed to enhance model generalization and interpretability \citep{prakashMoreRealisticSimulated2022,leeEvoAugImprovingGeneralization2023,seitzInterpretingCisRegulatory2023}. In \citet{prakashMoreRealisticSimulated2022}, sequence generation is guided by a motif-based pipeline to fine-tune trained models and benchmark different \textit{post-hoc} explainability methods. In contrast, EvoAug \citep{leeEvoAugImprovingGeneralization2023} applies random augmentations to the sequences during pretraining, using the same label as the wild-type sequence. Then, the potential biases and labeling errors created with the augmentations are addressed by fine-tuning on the original data only. The adoption of these augmentations has demonstrated improved generalization and more interpretable feature attribution maps. Finally, SQUID \citep{seitzInterpretingCisRegulatory2023} employs synthetic data generation to train interpretable surrogates to elucidate deep learning models in local regions of the sequence space. This approach aligns with our method, where oracle predictions serve as labels for the synthetic dataset. However, in SQUID, perturbations are applied randomly without ensuring comprehensive coverage of the model's semantic space. 
This might be a limitation, depending on the application. For instance, in our RNA splicing case study, if one wants to study what drives the splicing levels of a wild-type exon from 0\% inclusion to 100\%, it is unlikely that the randomly generated synthetic dataset will adequately cover the two far-reaching locations of the semantic space. 

\section{Proposed approach}
\label{sec:approach}

\subsection{Overview}
\label{subsec:approach}

We hypothesize that an evolutionary algorithm modeling the semantic perturbations of an input sequence guides the search for a good dataset better than randomly sampling from the search space. We propose a solution based on \gls{gp} to test this hypothesis. \gls{gp} explores the search space by maintaining a population of interesting programs or instances — in our case, combinations of perturbations — and by applying genetic operators such as mutation and crossover.   

\begin{figure*}
    \centering
    \includegraphics[width=0.9\textwidth]{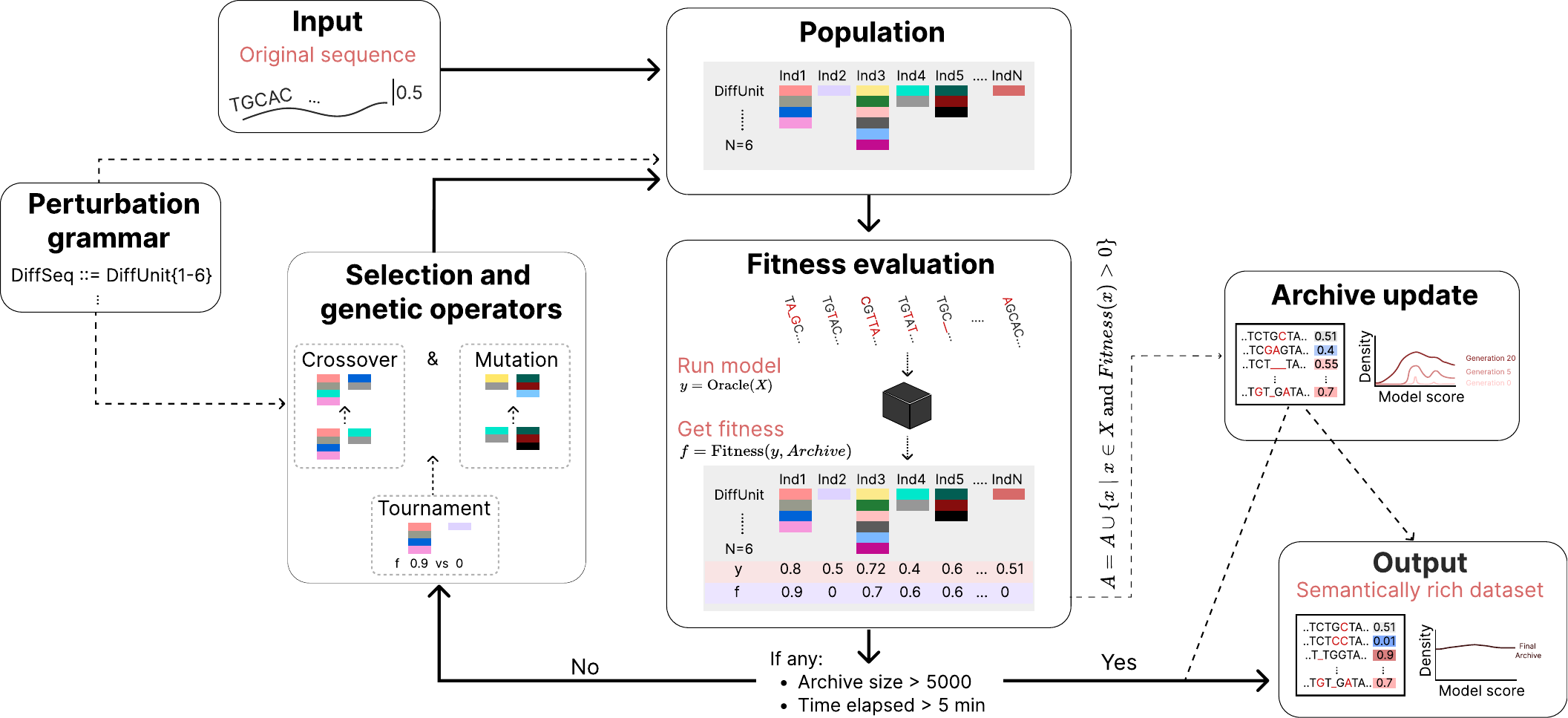}
    \caption{Summary of the proposed methodology.}
    \label{fig:evolutionary_method}
\end{figure*}

Throughout the evolution cycle, relevant individuals are copied from the population to an archive. Once the evolution ends, either due to a time limit or reaching the archive capacity, this archive becomes the final dataset. Thus, the population is used to keep individuals that represent promising areas of the search space, not the final dataset (\Cref{fig:evolutionary_method}).

We use a domain-specific representation for individuals. Rather than sequences as strings of nucleotides, we choose to represent the perturbations themselves using a grammar (\Cref{subsec:representation}). This representation draws inspiration from diff files, which compactly represent two similar files by listing only their differences. It offers several advantages, including reduced memory usage compared to storing two complete copies and being easier for practitioners to interpret. Additionally, genetic operators can work on the semantic level of a perturbation, allowing the introduction of domain-specific and biological constraints.

Consequently, the initial population is randomly generated using the semantic rules encoded in the grammar. This population undergoes evaluation by applying the perturbation representation to the original sequence, resulting in perturbed sequences. These sequences are then fed into the deep learning oracle to obtain predictions in the semantic space of the model. Our goal is not to optimize the prediction itself, but rather the diversity within the archive's predictions (\Cref{subsec:archive}). To achieve this, we define two fitness functions (\Cref{subsec:fitnessfunctions}) that are computationally lightweight and can guide the population towards archive diversity. Finally, we follow a traditional \gls{gp} loop, with tournament selection (size 5) and application of grammar-guided genetic operators such as tree-based crossover and mutation (\Cref{subsec:operators}). Additionally, we propose a custom mutation operator that promotes locality in the sampled positions of perturbations, acting as a more local search than the traditional tree-based mutation.

\subsection{Representation}
\label{subsec:representation}

We devised a perturbation grammar to constrain the perturbations to be biologically plausible. Using \gls{gggp}~\citep{whighamGrammaticallybasedGeneticProgramming1995}, extended with meta-handlers~\citep{espadaDataTypesMore2022}, both the population initialization and genetic operators modify the representation of individuals within these constraints.

\begin{figure}
\begin{bnfgrammar}
DiffSeq : Perturbation Sequence ::= DiffUnit\{1-6\}
;;
DiffUnit : Perturbation ::= SNV
    | Insertion
    | Deletion
;;
SNV : SNV  ::= \underline{SNV}(Pos, Nuc)
;;
Insertion : Insertion ::= \underline{Ins}(Pos, Nucs)
;;
Deletion : Deletion ::= \underline{Del}(Pos, Size)
;;
Nucs : Multiple Nucleotides ::= Nuc+
;;
Nuc : Nucleotide ::= \underline{A} | \underline{C} | \underline{G} | \underline{T}
;;
Pos : Position ::= \underline{int}
;;
Size : Size of deletion ::= \underline{int}
\end{bnfgrammar}
\caption{The core structure of the grammar used to represent an individual in respect to the original sequence, presented in EBNF. Underlined symbols are terminals or meta-handlers.}
\label{fig:grammar}
\end{figure}

Our grammar (\Cref{fig:grammar}) defines individual genotypes as a sequence of 1 to 6 perturbations (\texttt{DiffSeq}, the starting symbol). Each perturbation (\texttt{DiffUnit}) can be one of three types: a \gls{snv}(\texttt{Pos,Nuc}) parameterized with a position and a new nucleotide, an insertion of a short sequence at a given position (\texttt{Insertion(Pos,Nucs)}), or a deletion of $n$ nucleotides at a given position (\texttt{Deletion(Pos, Size)}). \Cref{tab:representation_examples} provides examples of different perturbations and their effect on the resulting sequence.

Four principles guided the design of grammar: Firstly, the number of perturbations was capped at 6 to preserve similarity to the original sequence. Secondly, the three types of \texttt{DiffUnit}s, despite their varying lengths, represent types of genetic variation found in nature. We used meta-handlers to restrict the length of deletions and insertions to a maximum of 5. Thirdly, we prevent overlapping perturbations as it is impractical to keep track of the correspondence between positions and the original sequence. In these cases, only the largest perturbation is kept, hence prioritizing insertions and deletions over \gls{snv}s. Finally, we impose additional constraints on exploring specific regions within the sequence based on problem-specific requirements. This is particularly relevant when exploring certain regions that could significantly impact the search process, potentially leading to local optima. For instance, in the context of the RNA splicing problem, we explicitly restrict any \texttt{DiffUnit} within the positions \texttt{[-10, 2]} and \texttt{[-3, 6]} around splicing acceptors (the start of an exon) and donors (the end of an exon), respectively. These restrictions are enforced using meta-handlers on the values of the \texttt{Pos} non-terminal.

Ultimately, the genotype is a list of non-overlapping perturbations applied to the original sequence.

\begin{table}[]
    \caption{Examples of perturbations and their effect on the original sequence.}
    \begin{tabular}{@{}cll@{}}
    \toprule
    \multicolumn{1}{l}{Original Sequence} & Perturbation              & Final Sequence \\ 
    \midrule
    \multirow{4}{*}{ATTCGCGTTA}           & {[}SNV(1,A){]}            & A{\color{purple}A}TCGCGTTA     \\
                                          & {[}Ins(2,CG){]}*          & AT{\color{purple}CG}TCGCGTTA   \\
                                          & {[}Del(7,2){]}*           & ATTCGCGA       \\
                                          & {[}SNV(1,A), Ins(2,CG){]} & A{\color{purple}A}{\color{purple}CG}TCGCGTTA  \\
    \bottomrule
    \end{tabular}
    \label{tab:representation_examples}
    \footnotesize{\parbox[t]{\linewidth}{* Ins stands for Insertion; Del stands for Deletion.}}
\end{table}

\subsection{Archive}
\label{subsec:archive}

The generation of the archive is the main outcome of the evolutionary algorithm. It is used as a dataset for any downstream application. The archive has a fixed specified capacity $S$ and is composed of $N$ equally-sized bins (or buckets). In this study, we used 5000 and 40 bins, each representing a range of 0.025 within the black box prediction space $\{ p \in \mathbb{R} \mid 0 \leq p \leq 1 \}$. The optimal archive would maintain an equal and maximum capacity in all bins while also displaying good diversity within each bin. 

The quality ($Q$) of an archive $A$ is the weighted sum of the number of sequences stored in the archive ($A_{size}$), the archive inter-bin diversity ($\hat{D}$), the intra-bin diversity ($\hat{D}_{per\_bin}$) and the fraction of bins with at least 10 sequences ($A_{no\_low\_count\_bins}$):

\begin{equation}
    \label{eq:archive_quality}
    \begin{split}
        Q(A) &= 0.3 \times A_{size} \\
        &+ 0.3 \times \hat{D} \\
        &+ 0.2 \times \hat{D}_{Per\_bin} \\
        &+ 0.2 \times (A_{no\_low\_count\_bins})
    \end{split}
\end{equation}

The raw archive diversity $D$ is quantified using the Shannon diversity index and normalized ($\hat{D}$) to scale between 0 and 1:

\begin{equation}
    \label{eq:shannon}
    \begin{split}
        D(A, N) &= -\sum_{b=1}^{N} p_{A_b} \ln(p_{A_b}) \\
        \hat{D}(A, N) &= \frac{D(A, N)}{ln(N)}
    \end{split}
\end{equation}

where $p_{A_b}$ stands for the proportion of the sequences in the archive belonging to the \emph{b\textsuperscript{th}} prediction bin.

The intra-bin diversity $D_{per\_bin}$ measures the average diversity within each of the $N$ bins by further dividing each archive bin ($A_b$) into 10 equally-sized sub-bins and calculating the diversity $\hat{D}(A_b, 10)$ for each. The final $\hat{D}_{per\_bin}$ is the average diversity across all bins:

\begin{equation}
    \label{eq:diversity_per_bin}
    \hat{D}_{per\_bin} = \frac{1}{N} \sum_{b=1}^{N} \hat{D}(A_b,10)
\end{equation}

Finally, the $A_{no\_low\_count\_bins}$ quantifies the fraction of bins with more than 10 sequences. These factors balance a semantic representation that is both coarse and fine-grained, ensuring an even distribution across all bins, and consider the total number of sequences in the final dataset. While other metrics could be considered, we chose these for their relevance to our case study.

\subsection{Fitness Functions}
\label{subsec:fitnessfunctions}

The purpose of the fitness function is to assess how likely an individual is to be kept in the next generation. For example, even the worst individual of a given generation can be added to the archive if it helps improve its quality (fitness $>$ 0). However, it will probably not survive for the next generation and its genotype will be lost.

We define two fitness functions that take into account the current archive status: \emph{Bin Filler} and \emph{\gls{iad}}.

Bin Filler: This fitness function is directly proportional to the number of available slots in the bin that the current individual $i$ belongs to. It is defined as one minus the ratio between the number of archive sequences in the bin $b$ ($p_{A_b}$) and the target number of sequences per bin $T$. $T$ is computed \emph{a priori}, based on the desired archive size $S$ and the number of bins $N$:

\begin{equation}
    \label{eq:bin_filler}
    \begin{split}
    BF_i &= 1 - \frac{p_{A_b}}{T} \\
    T &= \frac{S}{N}. 
    \end{split}
\end{equation}

It aims to promote the survival of individuals that belong to emptier bins. This enhances the exploration of a combination of perturbations in the least explored areas of the oracle semantic space.

\gls{iad}: This fitness function measures how much the addition of individual $i$ to the Archive increases its inter-bin diversity, as described in \Cref{eq:shannon}:

\begin{equation}
    \label{eq:iad}
    IAD_i = \hat{D}({A\cup i}, N) - \hat{D}(A, N).
\end{equation}

It is designed to be less reliant on the current filling of each bin. Instead, it assigns higher fitness to a sequence if it positively contributes to the overall archive uniformity at that moment. This strategy might be advantageous in avoiding being stuck on local optima since evolution favors sequences that fully deviate from them. Nevertheless, both fitness functions share the same overall goal.

\subsection{Genetic Operators}
\label{subsec:operators}

We use standard tree-based \gls{gggp} mutation and crossovers, extended with meta-handlers~\citep{espadaDataTypesMore2022}. In a typical \gls{gggp} mutation, a mutation at a given position of the list would generate random, new elements for the remainder of the list. Through the usage of meta-handlers, mutations on lists result in either adding, removing or replacing exactly one element.

We also designed a custom mutation operator that replaces a randomly chosen \texttt{DiffUnit} with another one in proximity. This replacement is determined by a normal distribution centered at the position of the old \texttt{DiffUnit}. This approach enhances the search for functional local motifs in the sequence, a known property of biological sequences. As an example, a typical mutation in a \gls{gggp} individual, like \texttt{[SNV(7, G)]}, could replace the node position $7$ with a randomly sampled integer such as $8345$. Our custom mutation would select the node and replace it with \texttt{[SNV($v$, G)]}, with  $v \sim N(7,4)$. 

\section{Evaluation methodology}
\label{sec:evaluation_methodology}

To assess our \gls{gggp} method for local dataset generation, we employ it to synthesize local datasets for explaining SpliceAI, a neural network that models RNA splicing (\Cref{subsec:casestudy}). We describe the specific experimental settings, including hardware and software details, in \Cref{subsec:experimental_setup}. Additionally, in \Cref{subsec:baseline}, we detail the baseline approach against which our methodology is compared. Finally, \Cref{subsec:hyperopt} details our process for tuning hyperparameters.

\subsection{Case Study}
\label{subsec:casestudy}

While several applications for sequence generation do exist, our evaluation focuses on the problem of RNA splicing. In particular, we aim at generating synthetic datasets for local explainability of the SpliceAI model~\citep{jaganathanPredictingSplicingPrimary2019}. SpliceAI is a deep residual network that predicts the probability of each position in an input sequence to be a splice site. Besides accurately predicting these exon boundaries, SpliceAI has shown remarkable success in the prediction of new pathogenic variants, with vast external evidence supporting such capability \citep{rowlandsComparisonSilicoStrategies2021,haPerformanceEvaluationSpliceAI2021,desainteagatheSpliceAIvisualFreeOnline2023}. This was particularly striking in regions of the genome that were historically difficult to predict \citep{lopesCrypticSpliceAlteringVariants2020,qianIdentificationDeepIntronicSplice2021,barbosaComputationalPredictionHuman2023a}, suggesting that the model has indeed learned, at least partially, mechanistic rules of the splicing code.

In this study, the input is a DNA sequence representing an exon triplet along with the intervening introns. The goal is to generate sequences that influence the probability of inclusion of the middle exon (the so-called cassette exon). In biological terms, we target the generation of sequences to model exon skipping, the most prevalent alternative splicing event in the human genome~\citep{garcia-perezLandscapeExpressionAlternative2023}. Although SpliceAI does not directly model \gls{psi}, a well-established metric for quantifying exon inclusion levels, we use the average of SpliceAI predictions at the acceptor and donor positions of the cassette exon as a proxy for \gls{psi} values. This decision is justified by the observed correlation with \gls{psi} measurements from RNA-Seq data~\citep{jaganathanPredictingSplicingPrimary2019}.

The SpliceAI input is bounded to 10,001 nucleotides, ensuring that 5k of flanking context on both sides is considered for prediction of each central position undergoing evaluation. Sequences (exon triplets) shorter than this resolution were padded whereas sequences longer than 10k nucleotides were trimmed to conform with the model input dimensions.

As a proof-of-concept, we used the exon 6 of the FAS gene, an exon extensively studied \citep{cascinoThreeFunctionalSoluble1995,izquierdoRegulationFasAlternative2005,julienCompleteLocalGenotypephenotype2016,baeza-centurionCombinatorialGeneticsReveals2019} due to the fact that excluding this exon switches the protein's function from pro-apoptotic (programmed cell death) to anti-apoptotic. In addition, the \gls{psi} levels of this exon vary across tissues and displayed intermediary \gls{psi} levels of 60\% in a minigene construct containing exons 5-7 and the corresponding introns~\citep{baeza-centurionCombinatorialGeneticsReveals2019}. Similarly, using the same genomic context, SpliceAI predicts a \gls{psi} value of 0.4921 for exon 6, aligning with the observed behavior in real cells.

\subsection{Experimental settings}
\label{subsec:experimental_setup}

Our approach and the baseline were implemented on top of GeneticEngine v0.8.5 \citep{espadaDataTypesMore2022}, which supports meta-handlers that allow encoding constraints on the perturbations. Specifically, we developed \gls{dress} v0.0.1 (\url{https://github.com/PedroBarbosa/dress}), which incorporates the proposed techniques (see Supplementary Information for details).

All the experiments were conducted on a Ubuntu 22.04 server with an AMD Ryzen Threadripper 3960X 24-Core Processor with 96GB of usable RAM. The GPU used for model inferences was an NVIDIA GeForce RTX 3090 with 24Gb of VRAM, running on CUDA v12.3 and Python 3.10.12. Reproducibility instructions are available at \url{https://github.com/PedroBarbosa/Synthetic_datasets_generation}. The datasets generated in this study are available on Zenodo at \url{https://doi.org/10.5281/zenodo.10607868}.

\subsection{Baseline}
\label{subsec:baseline}

Existing work that generates local synthetic sequences employs either random ~\citep{seitzInterpretingCisRegulatory2023} or exhaustive (from a short 500-nucleotide sequence ~\citep{taskiranCellTypeDirected2023}) sampling. Since exhaustive search is impractical for large search spaces like ours, we adopt Random Search as the baseline. It is worth noting that the baseline also takes advantage of our semantically rich encoding and does not operate on sequences directly, thus enabling us to focus the evaluation on the impact of the \gls{gp} loop.

Both approaches are compared with the same time budget and are implemented on the same framework, reducing the impact of external factors in our evaluation.

\subsection{Hyperparameter Optimization}
\label{subsec:hyperopt}

\begin{table*}
    \caption{List of hyperparameters tuned by Optuna along with the best values for each strategy}
    \label{tab:tuned_parameters}
    \footnotesize
    \begin{tabular}{cccccc}
      \toprule
      Hyperparameter & Search space & GGGP\_BF & GGGP\_IAD & RS\_BF & RS\_IAD \\
      \midrule
      Max \emph{DiffUnits} & Int\{1,2,3,4,5,6\} & 5& 4& 6& 5\\
      Max insertion size & Int\{1,2,3,4,5\} & 5& 5& 5& 5\\
      Max deletion size & Int\{1,2,3,4,5\} & 3& 1& 5& 4\\
      SNV grammar weight & Float[0, 1] (Step 0.05) & 0.05& 0.15& 0.1& 0.25\\
      Insertion grammar weight & Float[0, 1] (Step 0.05) & 0.75& 0.25& 0.85& 0.4\\
      Deletion grammar weight & Float[0, 1] (Step 0.05) & 0.3& 0.1& 0.15& 0.35\\
      \midrule
      Population size & Int[100, 1900] (Step 200) & 500& 700& 1300& 1900\\
      Selection method * & Cat\{Tournament, Lexicase\} & Tournament& Tournament& -& -\\
      Crossover probability * & Float[0.05, 5] (Step 0.05) & 0.25& 0.2& -& -\\
      Mutation probability * & Float[0.2, 1] (Step 0.1) & 0.7& 0.7& -& -\\
      Use custom mutation operator * & Bool\{True, False\} & True& True& -& -\\
      Custom mutation operator weight * & Float[0, 1] (Step 0.1) & 0.8& 0.7& -& -\\
      Genetic operators weight * & Float[0, 1] (Step 0.1) & 0.8& 0.8& 0& 0\\
      Elitism weight * & Float[0, 1] (Step 0.1) & 0& 0.1& 0& 0\\
      Novelty weight * & Float[0, 1] (Step 0.1) & 0.1& 0& 1& 1\\
    \bottomrule
  \end{tabular}
  \begin{tablenotes}
    \item * In Random Search, these hyperparameters were not optimized. We strictly set Novelty weight to 1 and Elitism and Genetic operators weights to 0, turning the other highlighted parameters untouched.
  \end{tablenotes}
\end{table*}

We used Optuna v3.4.0 \citep{akibaOptunaNextgenerationHyperparameter2019} for hyperparameter optimization of the evolutionary algorithm. The objective was to identify the optimal configuration that maximized archive quality, as in \Cref{eq:archive_quality}. We used \gls{tpe}~\citep{bergstraAlgorithmsHyperparameterOptimization2011} for parameter sampling. The optimization process was carried out until 500 trials were successfully completed. Each trial was set to finish when either of the following conditions was met: the archive accumulated 5000 sequences, or the allocated time budget of 5 minutes was reached. We also added soft constraints on the sum of certain hyperparameters, favoring their total to be between 0.5 and 1. These constraints were applied to three sets of parameters: the sum of \emph{SNV grammar weight}, \emph{Insertion grammar weight} and \emph{Deletion grammar weight}; the sum of \emph{Genetic operators weight}, \emph{Elitism weight} and \emph{Novelty weight}; and the sum of \emph{Mutation probability} and \emph{Crossover probability}. The search space for all hyperparameters is reported in Table~\ref{tab:tuned_parameters}. Individual optimizations were conducted for each of the fitness functions, resulting in four distinct optimization runs: \gls{gggp}\_BinFiller, RandomSearch\_BinFiller, \gls{gggp}\_\gls{iad} and RandomSearch\_\gls{iad}.

\section{Results}
\label{subsec:results}

\subsection{Performance comparison}
\label{subsubsec:perf_comparison}

\begin{figure}[t]
    \centering
    \begin{minipage}{0.25\textwidth}
        \centering
        \includegraphics[width=\linewidth]{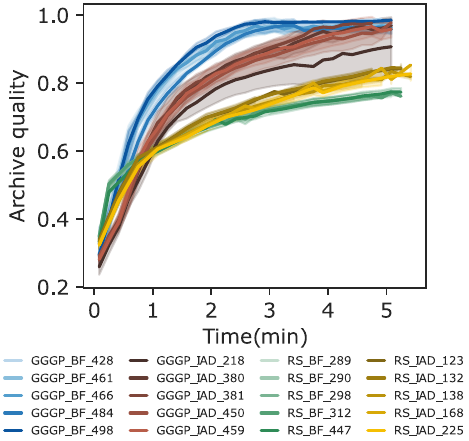}
    \end{minipage}\hfill
    \begin{minipage}{0.225\textwidth}
        \centering
        \includegraphics[width=\linewidth]{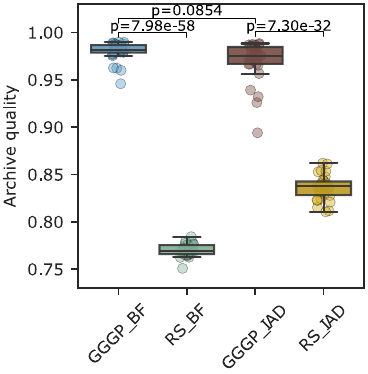}
    \end{minipage}
    \caption{Left: Average archive quality throughout the search procedure for four strategies (\gls{gggp}\_BinFiller, RandomSearch\_BinFiller, \gls{gggp}\_\gls{iad} and RandomSearch\_\gls{iad}). We show only the top 5 trials of each strategy, each representing the average of 5 seeds. Right: Distribution of the archive quality of the top trial of each strategy over 30 seeds. Statistical significance was assessed using Welch's t-tests for three pairs of samples: two comparing the means of the \gls{gggp} and baseline when using the same fitness function, and one comparing the \gls{gggp} with different fitness functions. P-values were adjusted for multiple testing using the Bonferroni correction.}
    \label{fig:perf_comparison}
\end{figure}

Firstly, we compared the top five trials of each of the four strategies throughout an evaluation using five seeds (\Cref{fig:perf_comparison}, left panel). We observed a high agreement across seeds within each strategy, except for a single trial with the \gls{iad} fitness function. The best overall configuration resulted from combining Genetic Programming with BinFiller, both in the final quality of the archive and the rate at which it increases throughout the evolution.

Next, we executed 30 seeds of the top trial of each strategy (\Cref{tab:tuned_parameters}). The larger number of seeds confirms the effectiveness of \gls{gggp} with BinFiller and highlights significant performance differences between \gls{gggp} and Random Search (\Cref{fig:perf_comparison}, right panel). When comparing the two fitness functions, \gls{gggp} with \gls{iad} was competitive against BinFiller, achieving a median archive quality of over 0.95. Interestingly, the hyperparameter search yielded a lack of novelty (weight 0 in \Cref{tab:tuned_parameters}), rendering the evolution highly dependent on the search space covered during population initialization. Additional experiments comparing fitness functions (see Supplementary Information for details) revealed a negligible effect on fitness function choice (Figure S1, left panel). However, the optimal parameter configuration for \gls{iad} exhibited lower average edit distances (Figure S1, right panel), which could be advantageous for downstream explainability applications—simpler genotypes may offer better interpretability.

\subsection{Ablation Studies}
\label{subsubsec:ablation_studies}

Using the best configuration, \gls{gggp} with BinFiller, we explored how different components can effect evolution performance.

\paragraph{Lexicase selection}
\label{lexicase}

\begin{figure}[t]
    \centering
    \begin{minipage}{0.23\textwidth}
        \centering
        \includegraphics[width=\linewidth]{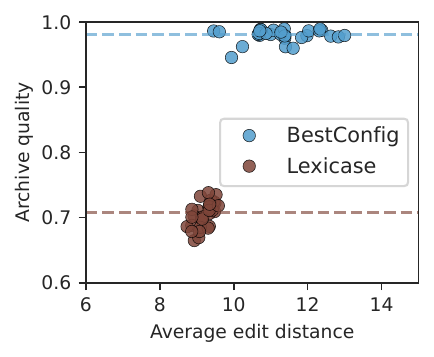}
    \end{minipage}\hfill
    \begin{minipage}{0.24\textwidth}
        \centering
        \includegraphics[width=\linewidth]{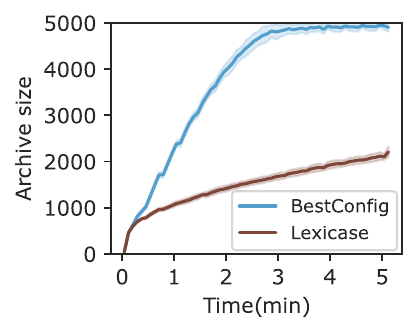} 
    \end{minipage}
    \caption{Left: Average edit distance and archive quality of tournament vs lexicase selection of 30 different runs. The horizontal lines reflect the median archive quality of each selection method. Right: Averaged archive size throughout 30 different runs.}
    \label{fig:lexicase}
\end{figure}

First, we replaced Tournament selection with Lexicase selection~\citep{spectorAssessmentProblemModality2012} with two objectives. Besides maximizing the quality of the archive (first objective), we also minimized the edit distance between the generated sequences and the original one. This second objective aims to reduce the syntactic diversity of the generated dataset, thereby potentially enhancing its explainability.

Surprisingly, Lexicase selection performed very poorly regarding archive quality (\Cref{fig:lexicase}, left panel). While the second objective helped the generation of simpler genotypes (lower average edit distance of the archives compared to the best configuration), this hurt overall performance. Custom hyperparameter tuning for Lexicase selection did not improve the results (data not shown). In addition, the slow rate at which sequences were added to the archive suggests that Lexicase hindered the exploration of sequence space (\Cref{fig:lexicase}, right panel). This is likely because most sequences added to the archive were selected based on the primary objective. Conversely, the second objective, which probably favored individuals with a single \gls{snv}, primarily slowed down the evolutionary process. This observation reflects biological complexity: higher edit distances, which are important for exploring epistatic interactions, are likely necessary for fully capturing the biological fitness landscape.

\paragraph{Custom mutation operator}
\label{custom_mutation_operator}

\begin{figure}[t]
    \centering
    \includegraphics[width=0.45\textwidth]{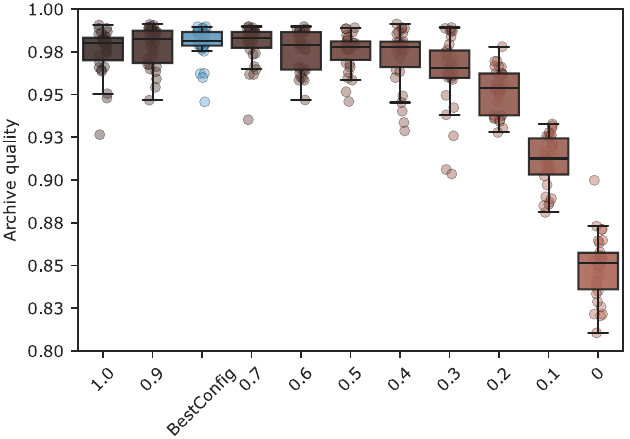}
    \caption{Impact of the frequency of the custom mutation operator (vs \gls{gggp} tree-based mutation) in the final archive quality, across 30 seeds.}
    \label{fig:custom_mutation}
\end{figure}

Next, we examined how the custom mutation operator affects archive quality. The best setup displayed a custom mutation rate of 0.8, where 80\% of \gls{gggp} mutations swap \texttt{DiffUnit}s with others in spatial proximity. We conducted experiments testing ten additional rates by incrementally decreasing this value from 1 to 0, consequently increasing the standard random mutation rate.

We found that reducing the custom mutation rate negatively impacted the overall quality of the archive, especially when relying solely on standard \gls{gggp} random mutation (rate 0), as illustrated in \Cref{fig:custom_mutation}. This outcome strongly suggests that exploring \texttt{DiffUnit}s in a more localized manner is targeting functional motifs faster than when mutating across the whole sequence. These results highlight the benefit of embedding domain and problem-specific properties in the design of the evolutionary algorithm.

\paragraph{Restricting the types of perturbations}
\label{grammar_nodes}

\begin{figure*}[b]
    \centering
    \begin{minipage}{0.39\textwidth}
        \centering
        \includegraphics[width=\linewidth]{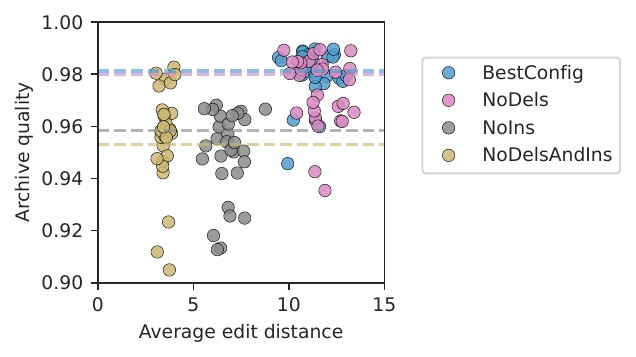}
    \end{minipage}\hfill
    \begin{minipage}{0.6\textwidth}
        \centering
        \includegraphics[width=\linewidth]{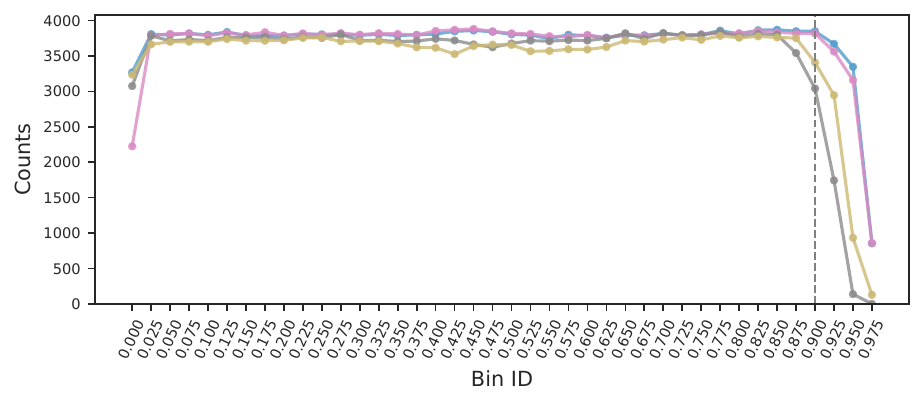} 
    \end{minipage}
    \caption{Impact of excluding specific grammar nodes on archive quality. Left: Average edit distance and archive quality for the best configuration, and the best configuration without deletions (\emph{NoDels}), without insertions (\emph{NoIns}) and without both (\emph{NoDelsAndIns}), each with 30 seeds. Horizontal lines represent the median archive quality for each condition. Right: Number of unique sequences generated across 30 seeds in each score bin for each condition. 
    }
    \label{fig:grammar_nodes}
\end{figure*}

Lastly, we assessed the performance impact of omitting certain \texttt{DiffUnits} from the grammar. As expected, when omitting both deletions and insertions (only \gls{snv}s are allowed) the archive quality was reduced, as there was no sufficient sequence edits to explore the whole model prediction landscape (Figure \ref{fig:grammar_nodes}, left panel, \emph{NoDelsAndIns} points).

Interestingly, when only insertions were excluded from the evolution, archive quality also dropped to similar levels as for the \emph{NoDelsAndIns} configuration. In contrast, the exclusion of deletions (\emph{NoDels}) appeared to have minimal impact on performance (Figure \ref{fig:grammar_nodes}, left panel). This pattern suggests that insertions play a crucial role in improving black box prediction coverage. In particular, this is evident for score bins greater than 0.9, which are the hardest to reach under all conditions: grammars without insertions (\emph{NoIns, NoDelsAndIns}) contribute disproportionately fewer sequences at these bins compared to grammars with insertions (\emph{NoDels, BestConfig}, as shown in Figure \ref{fig:grammar_nodes}, right panel).

\subsection{Generalization}

\begin{figure}[h]
	\centering
    \includegraphics[width=1\linewidth]{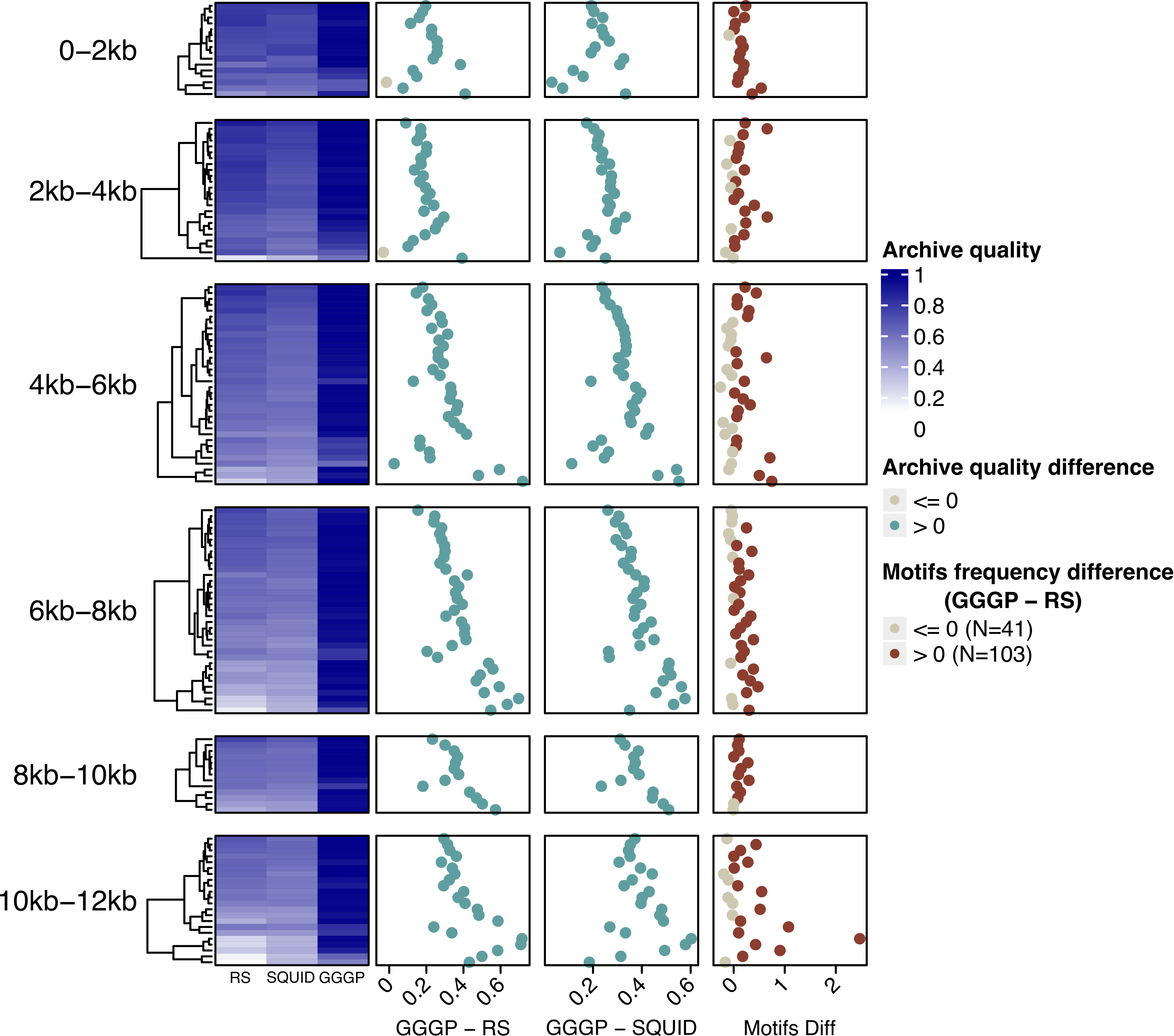}
    \caption{Heatmap comparing archive qualities across sequences of varying lengths (binned vertically), over five different seeds. The additional annotations, in green, show the performance difference between \gls{gggp} and each of the baselines. The right-most heatmap annotation summarizes motif analysis by displaying the difference in the relative frequency of motif disruptions between \gls{gggp} and Random Search.}
    \label{fig:generalization1}
\end{figure}

\begin{figure}[h]
    \centering
	\includegraphics[width=0.75\linewidth]{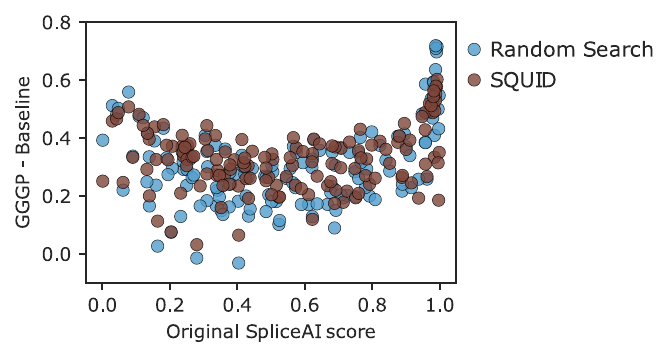} 
    \caption{Differences in archive quality between \gls{gggp} and the baselines (Random Search and SQUID) as a function of the SpliceAI score of the original sequence.}
    \label{fig:generalization2}
\end{figure}

To assess the generalizability of our approach to other sequences, we reanalyzed \gls{rnaseq} data from the ENCODE \citep{vannostrandLargescaleBindingFunctional2020}, specifically focusing on the \gls{kd} of the RBFOX2 gene, a known regulator of alternative splicing \citep{gehmanSplicingRegulatorRbfox22012,venablesRBFOX2ImportantRegulator2013,jbaraRBFOX2ModulatesMetastatic2023}. Our goal was to globally assess the impact of the \gls{kd} and to identify exons potentially regulated, either directly or indirectly, by RBFOX2. We found 144 exons that were sensitive to the \gls{kd} (details in the Supplementary Information), which were subjected to analysis using our \gls{gggp} approach and the Random Search baseline. We have also integrated SQUID \citep{seitzInterpretingCisRegulatory2023} into the benchmarks as a second baseline (see Supplementary Information for details).

We found that \gls{gggp} outperformed Random Search and SQUID in dataset generation quality across a diverse set of input sequences (\Cref{fig:generalization1}). On average, \gls{gggp} achieved 0.93 in archive quality, compared to Random Search's 0.62 and SQUID's 0.60, representing a $\approx$30\% improvement. Random Search outperformed \gls{gggp} in only 2 out of 144 sequences (0.01\%), and even then, the margin was minimal. The difference in \gls{gggp} performance was more pronounced in longer sequences, underscoring that our approach better navigates larger search spaces (\Cref{fig:generalization1}, GGGP-RS and GGGP-SQUID heatmap annotations). Notably, \gls{gggp} performance appears unaffected by sequence length, maintaining consistent archive quality across different sequence sizes (\Cref{fig:generalization1}, Table S1).

We further illustrate the influence of the original sequence prediction on the search outcome. In particular, when the model predicts values close to 0 or 1, we observe larger performance differences between our approach and the baselines, especially when the original exon is predicted with high probability (\Cref{fig:generalization2}). This makes biologically sense because exons predicted to be constitutive (close to 1) or barely included (close to 0) in the final RNA transcript are inherently resistant to changes across the \gls{psi} landscape \citep{baeza-centurionMutationsPrimarilyAlter2020}. These findings underscore the effectiveness of our approach, especially considering our deliberate avoidance of perturbing highly sensitive regions in the sequences, such as splice sites.

Finally, we explored which known biological motifs were captured by the generated datasets. We quantified the frequency of what we term 
``motif disruption'' events (motif gains/losses relative to the original sequence) in datasets generated by \gls{gggp} and Random Search (see Supplementary Information for details). Our analysis revealed a higher proportion of motif disruption events in \gls{gggp} datasets (\Cref{fig:generalization1}, Motif Diff heatmap annotation), indicating that sequences generated by \gls{gggp} contain richer biological information. However, 41 exons (28.5\%) had datasets with a lower number of disruption events (relative to the dataset size) compared to those generated by Random Search. As \gls{gggp} datasets effectively cover the prediction landscape, it remains unclear whether the model has learned previously unknown motif syntax or if the perturbations influencing the model's predictions are merely spurious artifacts.

\section{Conclusions and Future Work}
\label{sec:conclusion}

Based on the accumulated evidence, we conclude that our approach greatly improves over random sampling on the task of generating semantically meaningful local synthetic datasets. We found this to be true not only in a relatively short, controlled sequence but also for 144 sequences that are diverse in their length, genomic location, and original black box score. In these sequences, our approach achieved an average 30\% improvement over the baseline.

Our results have also highlighted the advantage of introducing domain knowledge in the problem specification. By constraining highly sensitive regions in the sequence from being explored, we force the evolutionary algorithm to learn alternative yet biologically interesting paths to achieve semantic diversity. Furthermore, our custom mutation operator that promotes locality proved beneficial compared to \gls{gggp}'s default tree-based mutation.

In this study, our primary focus was on evaluating datasets using metrics based on the evolutionary algorithm we purposed. In future work, we aim to apply different attribution-based explainability methods on these synthetic datasets and assess their ability to identify ground-truth biological motifs. Additionally, we plan to introduce a motif-based grammar to replace random insertions and deletions with more fine-grained biological constraints. These enhancements will contribute to a comprehensive framework for conducting cost-effective \textit{in silico} experiments aimed at studying RNA splicing regulation.

\subsection*{Limitations}

Our study relied on SpliceAI as the oracle for all experiments. It's important to note that this model lacks tissue or cell-type specificity. While SpliceAI is invaluable for studying general RNA splicing mechanisms, it isn't suited for generating datasets tailored to specific cell contexts. We attempted to use Pangolin \citep{zengPredictingRNASplicing2022}, a tissue-specific model, but its inference time was too high for the purpose of this study.

\begin{acks}
    This work was supported by FCT through a fellowship to P.B., ref. SFRH/BD/137062/2018, project RAP, ref.EXPL/CCI-COM/1306/2021 (\url{https://doi.org/10.54499/EXPL/CCI-COM/1306/2021}), project HPC, ref. 2022.15800.CPCA.A1, and the LASIGE Research Unit, ref. UIDB/ 00408/2020 (\url{https://doi.org/10.54499/UIDB/00408/2020}) and ref. \\UIDP/00408/2020 (\url{https://doi.org/10.54499/UIDP/00408/2020}).
\end{acks}
\bibliographystyle{ACM-Reference-Format}
\bibliography{refs}

\end{document}


\section*{Supporting information for ``Semantically Rich Local Dataset Generation for Explainable AI in Genomics"}

\subsection*{Comparing fitness functions}

We optimized the hyperparameters for each evolutionary strategy, as outlined in Table 1 of the main manuscript. To further explore the impact of the fitness functions, we conducted a small experiment wherein we used the best parameters derived by Optuna for \gls{gggp} with \gls{iad} and BinFiller (BF), but exchanged the fitness functions between them. The best results stemmed from the parameter configuration derived for BinFiller, rather than from the specific fitness function used (Figure~\ref{suppfig:check_ff}, left panel). This suggests that the fitness function may not be the primary driver of the evolutionary search outcome; instead, the parameter setup appears to play a major role. Interestingly, when examining the average edit distances to the original sequence, we observed that the parameter setup obtained for \gls{iad} (IAD, BF\_IADsetup) yielded simpler genotypes, regardless of the fitness function (Figure~\ref{suppfig:check_ff}, right panel). Hence, although not being the most performant in archive quality, this configuration could prove beneficial for downstream interpretability applications.
Four strategies are depicted: 
\begin{figure}[ht]
  \centering
  \includegraphics[width=0.9\linewidth]{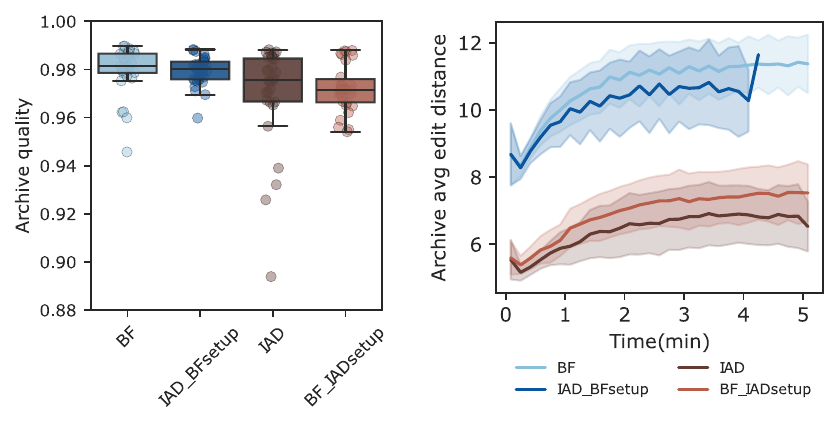}
  \caption{Evaluating the effect of the fitness function in the evolutionary search outcome. Four different strategies are displayed: Optuna-derived \gls{gggp} parameters for BinFiller and IAD, plus two experiments. These involve using IAD as fitness function with the BinFiller-optimized parameters (IAD\_BFsetup), and vice versa (BF\_IADsetup). Left plot: boxplot displaying archive quality variability based on 30 different seeds. Right plot: lineplot showing the average edit distance (number of nucleotides that differ from the original sequence) of the archive sequences along the evolution time based on 30 different seeds.}
  \label{suppfig:check_ff}
\end{figure}  

\subsection*{Generalization}

\subsubsection*{Differential splicing of RBFOX2 Knockdown data}

We obtained HepG2 RNA-Seq genome alignment files from ENCODE \cite{vannostrandLargescaleBindingFunctional2020}, specifically targeting the shRNA knockdown of RBFOX2 (experiment \href{https://www.encodeproject.org/experiments/ENCSR767LLP/}{ENCSR767LLP}). Similarly, control experiments were downloaded (experiment \href{https://www.encodeproject.org/experiments/ENCSR104ABF/}{ENCSR104ABF}). We employed rMATS v4.1.2 \cite{shenRMATSRobustFlexible2014} using GENCODE annotations v44 \cite{frankishGENCODE20212021} to detect differentially spliced events between control and knockdown replicates (\texttt{--cstat 0.05}). Significant events were defined by an absolute dPSI $>$ 0.2 with an FDR cutoff of 0.05. In addition, we applied a read coverage filter where we retained the events where the median coverage across replicates per condition for the isoform with more reads was higher than 5. Moreover, we exclusively focused on exon skipping events and removed events that shared the start or end coordinate with any other differentially spliced exon. This analysis resulted in a final set of 144 exons to be subjected to evolutionary searches.

\subsubsection*{Adding SQUID to the comparison}

We included SQUID v0.4.0 \cite{seitzInterpretingCisRegulatory2023} as an additional baseline for the generalization experiments. Importantly, SQUID's primary goal is not the generation of the dataset per se; instead, it focuses on using the generated dataset to train surrogate models to interpret genomics deep neural networks. At the dataset generation step, SQUID applies random perturbations to the input sequence. However, it differs from our Random Search approach in several aspects. Firstly, the perturbations are applied directly to the input sequence agnostically to our target specifications - generating diversity in the model prediction space - and thus are much faster to run. Secondly, SQUID restricts perturbations to SNVs and does not allow deletions and deletions. Lastly, it does not allow domain-aware constraints to be applied, such as avoiding perturbations in splice site regions. It was precisely due to this last point that we included SQUID in the comparison: to assess whether allowing perturbations at splice site regions (resulting in drastic changes) would enhance the coverage of the model prediction space.

To make SQUID comparable to our evolutionary-based approaches, we customized a SpliceAI predictor such that the predictions returned are the average of the model at the splicing acceptor and donor positions of the cassette exon. Then, for each input sequence we set up a \texttt{RandomMutagenesis} object with a \texttt{mut\_rate} value that yielded an average number of perturbations per sequence similar to the average edit distance of the datasets obtained for the best \gls{gggp} configuration (Figure 3 of the main manuscript, right panel), which was 11.3. Then, we generated 5000 sequences (equivalent to the archive capacity of the evolutionary-based approaches) using 5 different seeds. Finally, we converted the generated sequences into DRESS-compatible format and evaluated the quality of the datasets accordingly. Of note, we did not run the motif analyses for the SQUID-generated datasets because the perturbation space is different from Random Search and \gls{gggp} (splice site regions perturbed), rendering the comparison unfair.

\subsubsection*{Performance across different input sequence sizes}

We aimed to assess the impact of the sequence length on performance. Since the proof-of-concept was carried out on a relatively short sequence (FAS exon triplet plus 100 nucleotides on each side, totaling 1743 nucleotides), we examined how the generated archives stand up to much longer sequences, which reflect larger search spaces. Remarkably, we found that our approach is resilient to variations in input sequence length, with archive qualities consistently remaining high, in contrast to Random Search and SQUID (Table \ref{tab:perf_per_seq_len_bin}).

\begin{table*}
  \centering
  \caption{Mean archive quality for sequences at different sequence length intervals}
  \label{tab:perf_per_seq_len_bin}
  \footnotesize
  \begin{tabular}{ccccc}
    \toprule
    Sequence length class & Number of sequences & Random Search & SQUID & GGGP \\
    \midrule
    0-2kb & 16 & 0.707 & 0.702 & 0.91 \\ 
    2kb-4kb & 24 & 0.732 & 0.678 & 0.917 \\ 
    4kb-6kb & 34 & 0.639 & 0.603 & 0.929 \\ 
    6kb-8kb & 35 & 0.562 & 0.552 & 0.943 \\ 
    8kb-10kb & 13 & 0.585 & 0.573 & 0.955 \\ 
    10kb-12kb & 22 & 0.494 & 0.514 & 0.909 \\ 
    \bottomrule
  \end{tabular}
\end{table*}


\subsubsection*{Motif analysis}

We performed a classical motif analysis on the generated datasets to assess the presence of ground-truth motifs in the synthetic sequences. We scanned motif \gls{pwms} across sequences using FIMO \cite{grantFIMOScanningOccurrences2011} from the MEME suite v5.5.3. We used oRNAment \cite{benoitbouvretteORNAmentDatabasePutative2020} as the PWM database, and selected \gls{rbps} associated with RNA splicing regulation (according to \cite{vannostrandLargescaleBindingFunctional2020}), which resulted in PWMs from 36 \gls{rbps} for scanning. Only motif matches with a p-value $< 0.0001$ were considered.

From the motif scanning step, we additionally processed the output to measure the frequency of motif disruption events per dataset, defined as the total number of motifs gained or lost relative to the original wild-type sequence, divided by the dataset size. The dataset size is just a normalization factor, since the number of generated sequences was not consistent between Random Search and \gls{gggp}-based datasets.

\subsection*{DRESS software}

We developed a software called \gls{dress} to ease the generation of local synthetic datasets for any given input sequence (exon). By providing a simple client interface, we expect \gls{dress} to be useful for computational researchers interested on using Deep Learning to study mechanisms of RNA splicing regulation. As for now, it has support for SpliceAI \cite{jaganathanPredictingSplicingPrimary2019} and Pangolin \cite{zengPredictingRNASplicing2022}, two of the most widely used splicing sequence-based models.

While for the scope of this manuscript the main application is the generation of local datasets that span model prediction landscape (e.g, archives with high diversity), \gls{dress} can be easily applied for some sort of sequence design, that is, generating sequences with desired splicing outcomes. For example, \gls{dress} supports archive filtering, which allows the retrieval of synthetic sequences at desired \gls{psi} levels or splice site probabilities. In addition, it can be easily extended with different fitness functions that guide the genetic algorithm into desired \gls{psi} distributions. An important note, however, is that running \gls{dress} is impractical on environments without GPU availability. Due to its reliance on black box model inferences during population evaluation, running \gls{dress} on a CPU would result in significantly slower execution. 

The software incorporates valuable domain knowledge, particularly beneficial for those familiar with genomics analysis:

\begin{itemize}
    \item Streamlines the whole preprocessing of the input by allowing common bioinformatics formats (e.g., BED files) and extracting critical genomic intervals for the target sequence (e.g, surrounding exons and splice site positions) based on an internal GTF exon cache.
    \item Managing grammar-based search space constraints is straightforward. By default, \gls{dress} avoids perturbing positions near splice sites to prevent drastic outcomes. However, users can control these ranges via additional input arguments. Furthermore, domain-informative tags can be provided to restrict the search space further. For instance, to tweak splicing without interfering with the cassette exon, one can run \gls{dress} with \texttt{--untouched\_regions cassette\_exon}.
    \item The software has an alpha version of an evaluation framework that assesses generated datasets at various levels. This includes examining sequence motifs or conducting phenotype-based analyses of enriched perturbed regions, thus aiding with the explainability of the black box model.
\end{itemize}

\gls{dress} is available at \url{https://github.com/PedroBarbosa/dress}, and we are working towards providing comprehensive documentation, tutorials, and specific use cases in the near future.

\bibliographystyle{ACM-Reference-Format}
\bibliography{refs}